\documentclass{article}

\usepackage{microtype}
\usepackage{graphicx}
\usepackage{bm}
\usepackage{booktabs} 
\usepackage{enumitem}

\usepackage{upgreek}
\usepackage{color, colortbl}
\usepackage[most]{tcolorbox}
\usepackage{caption}
\usepackage{subcaption}
\usepackage{multirow}
\usepackage{url}

\usepackage{amsmath,amssymb,amsfonts}
\usepackage{placeins}
\usepackage{algorithmic}
\usepackage{graphicx}
\usepackage{textcomp}
\usepackage{xcolor}
\usepackage{algorithmic}
\usepackage{graphicx}
\usepackage{textcomp}
\usepackage{adjustbox}
\usepackage{float}
\usepackage{svg}
\usepackage{mathrsfs} 
\usepackage{bm}
\usepackage{upgreek}
\usepackage{color, colortbl}
\usepackage[most]{tcolorbox}
\usepackage{caption}
\usepackage{subcaption}
\usepackage{multirow}
\usepackage{url}

\usepackage{icml2023}

\usepackage[capitalize,noabbrev]{cleveref}

\usepackage{icml2023}

\usepackage{amsmath}
\usepackage{amssymb}
\usepackage{mathtools}
\usepackage{amsthm}

\usepackage[capitalize,noabbrev]{cleveref}

\theoremstyle{plain}

\theoremstyle{definition}

\theoremstyle{remark}

\usepackage[textsize=tiny]{todonotes}

\icmltitlerunning{Rethinking Robust Contrastive Learning
from the Adversarial Perspective}

\begin{document}

\twocolumn[
\icmltitle{Rethinking Robust Contrastive Learning  \\         from the Adversarial Perspective}



\icmlsetsymbol{equal}{*}

\begin{icmlauthorlist}
\icmlauthor{Firstname1 Lastname1}{equal,yyy}
\icmlauthor{Firstname2 Lastname2}{equal,yyy,comp}
\icmlauthor{Firstname3 Lastname3}{comp}
\icmlauthor{Firstname4 Lastname4}{sch}
\icmlauthor{Firstname5 Lastname5}{yyy}
\icmlauthor{Firstname6 Lastname6}{sch,yyy,comp}
\icmlauthor{Firstname7 Lastname7}{comp}
\icmlauthor{Firstname8 Lastname8}{sch}
\icmlauthor{Firstname8 Lastname8}{yyy,comp}
\end{icmlauthorlist}

\icmlaffiliation{yyy}{Department of XXX, University of YYY, Location, Country}
\icmlaffiliation{comp}{Company Name, Location, Country}
\icmlaffiliation{sch}{School of ZZZ, Institute of WWW, Location, Country}

\icmlcorrespondingauthor{Firstname1 Lastname1}{first1.last1@xxx.edu}
\icmlcorrespondingauthor{Firstname2 Lastname2}{first2.last2@www.uk}

\icmlkeywords{Machine Learning, ICML}

\vskip 0.3in
]



\printAffiliationsAndNotice{\icmlEqualContribution} 




\begin{abstract}
To advance the understanding of robust deep learning, we delve into the effects of adversarial training on self-supervised and supervised contrastive learning alongside supervised learning. Our analysis uncovers significant disparities between adversarial and clean representations in standard-trained networks across various learning algorithms. Remarkably, adversarial training mitigates these disparities and fosters the convergence of representations toward a universal set, regardless of the learning scheme used. Additionally, increasing the similarity between adversarial and clean representations, particularly near the end of the network, enhances network robustness. These findings offer valuable insights for designing and training effective and robust deep learning networks. Our code is released at \textcolor{magenta}{\url{https://github.com/softsys4ai/CL-Robustness}}.

\end{abstract}

\section{Introduction}
Self-supervised learning has significantly improved in recent years, leading to state-of-the-art performance in various applications. While this paves the way to learn effective representations from massively available unlabeled data, the vulnerability to adversarial attack is still a fatal threat. Adversarial training is proven to be an effective defense method in supervised learning. This method can be interpreted as a min-max optimization problem \cite{madry2017towards}, wherein the model parameters are updated iteratively by minimizing a training loss against the adversarial perturbations generated by maximizing an adversary loss function. While it is standard practice to use the same loss function for both training and generating adversarial attacks, some works have explored the use of dissimilar loss functions to investigate robust training \cite{pal2021adversarial}.  Hendrycks et al. \cite{hendrycks2019using} introduced a self-supervised term into the training loss to improve the robustness of a supervised model.  Chen et al. \cite{chen2020adversarial} were the first to apply adversarial training on a self-supervised model to achieve robust pre-trained encoders that can be used for downstream tasks through fine-tuning.

In recent years, the study of adversarial training on the robustness of various contrastive learning schemes has attracted great attention. The main idea of Contrastive Learning (CL) is to benefit from comparing semantically similar against dissimilar samples to learn the proper representations.  Some recent works employed the generated adversarial examples as a similar match of the anchor data point to improve the robustness of the model. Kim et al. \cite{kim2020adversarial} were the first to utilize the contrastive loss to generate adversarial examples without any label for robustifying SimCLR \cite{chen2020simple} framework. Moshavash et al. \cite{moshavash2021momentum}, Wahed et al. \cite{wahed2022adversarial}, and Gowal et al. \cite{gowal2021self} have applied the same technique to Momentum Contrast (MOCO) \cite{he2020momentum}, Swapping Assignments between Views (SwAV) \cite{caron2020unsupervised} and Bootstrap Your Own Latents (BYOL) \cite{grill2020bootstrap}, respectively. Fan et al. \cite{fan2021does} introduced an additional regularization term in contrastive loss to enhance cross-task robustness transferability. They use a pseudo-label generation technique to avoid using labels in adversarial training of downstream tasks. Similarly, Jiang et al. \cite{jiang2020robust} considered a linear combination of two contrastive loss functions to study the robustness\footnote{We use the word ``robust" as shorthand for ``adversarially robust" throughout the paper.} under different pair selection scenarios.

One of the central steps in any contrastive learning scheme is the selection of positive and negative pairs. Without label information, a positive pair is often obtained by data augmentation, while the negative samples are randomly chosen from a mini-batch. However, this random selection strategy can lead to choosing the false-negative pairs when two samples are taken from the same class. In \cite{gupta2023contrastive}, Gupta et al. empirically demonstrate that the adversarial vulnerability of contrastive learning is related to employing false negative pairs during training. One remedy to this dilemma is to leverage the label information to extend the self-supervised contrastive loss into a supervised contrastive (SupCon) loss introduced by \cite{khosla2020supervised}. SupCon loss contrasts embeddings from the same class as positive samples against embeddings from different classes as negative samples. It has shown that SupCon outperforms cross-entropy loss in terms of accuracy and hyper-parameters stability \cite{khosla2020supervised}. Islam et al. \cite{islam2021broad} have conducted a broad study to compare the transferability of learned representations by cross-entropy, SupCon, and standard contrastive losses for several downstream tasks. Zhong et al. \cite{zhong2022self} have designed a series of robustness tests, including data corruptions, ranging from pixel-level gamma distortion to patch-level shuffling and dataset-level distribution shift to quantify differences between contrastive learning and supervised learning frameworks.

Some very recent literature has started explaining and understanding robust networks. Jones et al. \cite{jones2022if} have shown that irrespective of architecture or random initialization, adversarial robustness is a significant constraint on the learned function of a network. The research conducted by \cite{cianfarani2022understanding} provides insights into the effects of adversarial training on representations. It emphasizes the lack of specialization in robust representations and the significant impact of overfitting on deeper layers during robust training. However, it is important to note that the primary focus of this study is on supervised learning algorithms. To address this research gap, our study investigates the effect of robust training on models trained using different learning schemes (contrastive, supervised-contrastive, and supervised) through a unified lens.

\noindent \textbf{Contributions}
In this work, we conduct several comprehensive experiments to compare the robustness of contrastive and supervised contrastive with standard supervised learning under different training scenarios. Our research utilizes explanatory tools such as CKA (Centered Kernel Alignment) \cite{kornblith2019similarity, nguyen2020wide,subramanian2021torchcka} and linear probing \cite{alain2016understanding} to investigate and contrast the layer-wise representations learned through various training methods. The design and implementation of the experiments are motivated by the following research questions:
\vspace{-1em}
\begin{itemize}
    \item [Q1:] \textit{Is there anything special about the learned representation with contrastive learning in terms of adversarial robustness?}
    \item [Q2:] \textit{To what extent does employing the label information benefit or deteriorate the robustness of contrastive learning representations?}
    \item [Q3:] \textit{How does adversarial training affect (similarities and differences) the learned representations in supervised and contrastive learning?}
\end{itemize}

Our key findings can be summarized as follows:
\vspace{-1em}
\begin{itemize}
    \item [R1:] Our results show that \textbf{contrastive learning without label information is less robust than other learning schemes in standard training}. However, combining the standard contrastive loss with either supervised cross-entropy or supervised contrastive loss can improve the robustness of the learned representations by leveraging the label information. (Section \ref{Sec: EX1})
    \item [R2:] From our results, we can observe the \textbf{significant positive impact of full adversarial fine-tuning on the robustness of representations learned by contrastive learning}. However, full adversarial fine-tuning is ineffective in supervised contrastive or standard supervised learning schemes. (Section \ref{Sec: EX2})
    \item [R3:] Our study reveals important insights regarding the impact of adversarial training on representations in different learning schemes. We observed \textbf{substantial differences between adversarial and clean representations in standard-trained networks across various learning schemes}. However, \textbf{after adversarial training, we observed a remarkable similarity between adversarial and clean representations}. This indicates that \textbf{regardless of the learning scheme utilized, adversarial training facilitates the convergence of representations towards a universal set, characterized by features\footnote{A representation consists of all the features found in a layer.} that consistently emerge across different models and tasks} \cite{olah2020zoom}. Additionally, we found that \textbf{increasing the similarity between adversarial and clean representations, especially at the end of the network, improves the robustness of the network}. These findings offer valuable insights into designing and training more efficient and effective robust networks. (Section \ref{Sec: EX3})
\end{itemize}

\begin{figure}[hbt!]
     \centering
     \begin{subfigure}[b]{0.45\textwidth}
         \centering
         \includegraphics[width=\textwidth]{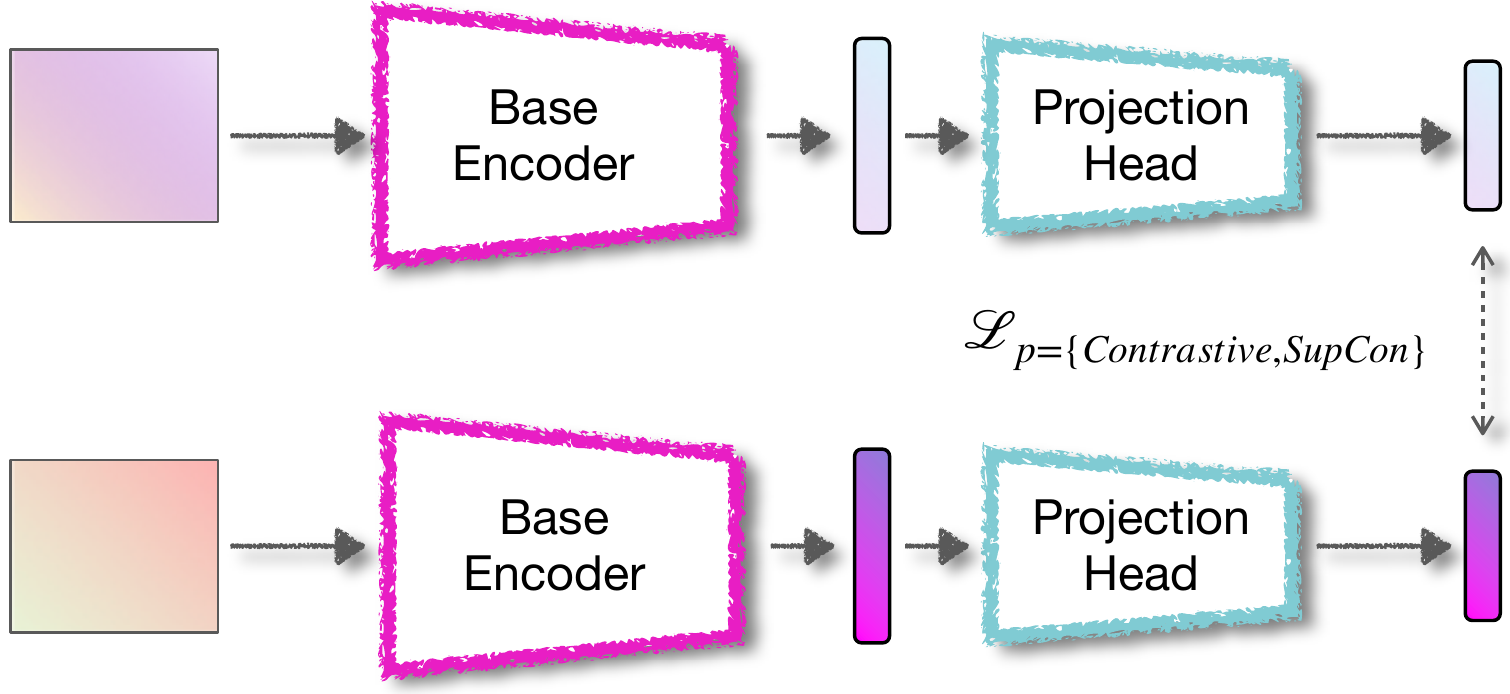}
         \caption{Contrastive and supervised contrastive learning.}
         \label{fig:contrastive}
     \end{subfigure}
     \hfill
     \begin{subfigure}[b]{0.45\textwidth}
         \centering
         \includegraphics[width=\textwidth]{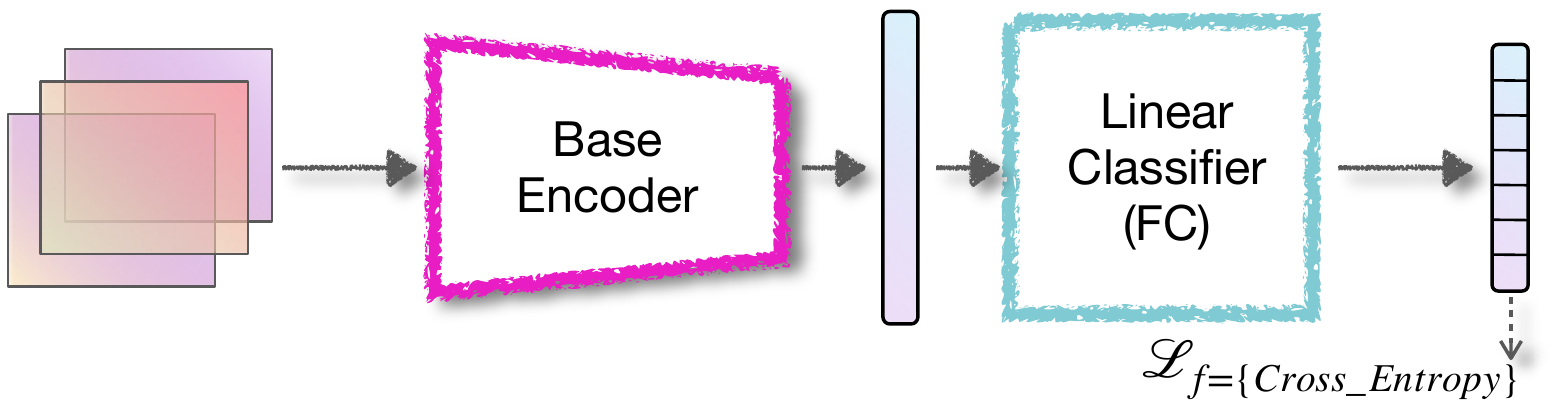}
         \caption{Supervised learning.}
         \label{fig:supervised}
     \end{subfigure}
        \caption{Training process of the studied learning schemes.}
        \label{fig:methodology}
\end{figure}

\section{Methodology}
In this section, we explain the methodology of our comparative study on the robustness of the three following learning schemes:

\begin{itemize}

\item \textit{Contrastive Learning (CL):} In the standard framework of SimCLR, contrastive learning trains a base encoder by minimizing a contrastive loss over the representations projected into a latent space (Figure \ref{fig:contrastive}). The extracted features will train a linear classifier on a downstream task.  

\item \textit{Supervised Contrastive Learning (SCL):} A supervised extension of contrastive learning introduced in \cite{khosla2020supervised}, to avoid false positive pairs selection by leveraging the label information. 
    
\item \textit{Supervised Learning (SL):} The network consists of a base encoder followed by a fully connected layer as a linear classifier (see Figure \ref{fig:supervised}). In this case, cross-entropy between the true and predicted labels is utilized for training the network parameters.
\end{itemize}

The training process in contrastive and supervised contrastive learning includes the following two phases:

\noindent\textbf{Pretraining Phase:} The goal of this phase is to train the base encoder parameters $\bm{\theta}_{b}$ by minimizing a self-supervised loss $\mathscr{L}_{p}(\bm{\theta}_{b},\bm{\theta}_{ph})$ over a given dataset $\mathscr{D}_{p}$. Here $\bm{\theta}_{ph}$ is the parameters vector of the projection head used to map the base encoder output into a low dimensional latent space where the $\mathscr{L}_{p}$ is applied.

\noindent\textbf{Supervised Fine-tuning Phase:} 
The goal of this phase is to train the linear classifier parameters $\bm{\theta}_{c}$ by minimizing the supervised loss $\mathscr{L}_f(\bm{\theta}_{c})$ over a labeled dataset $\mathscr{D}_f$.
The linear classifier learns to map the representations extracted during the pretraining phase to the labeled space, where $\mathscr{L}_f$ is the cross-entropy loss (See Appendix \ref{ap-methodology} for complete details of our methodology).
\\
We examine the standard and robust training variations of the aforementioned training phases to compare the adversarial robustness across different learning schemes. Table \ref{Ap.pf} in Appendix \ref{ap-methodology} summarises all the studied training combinations for different possible scenarios of training phases in contrastive and supervised contrastive learning schemes.

\begin{figure}[t]
\centering
\includegraphics[scale=0.35]{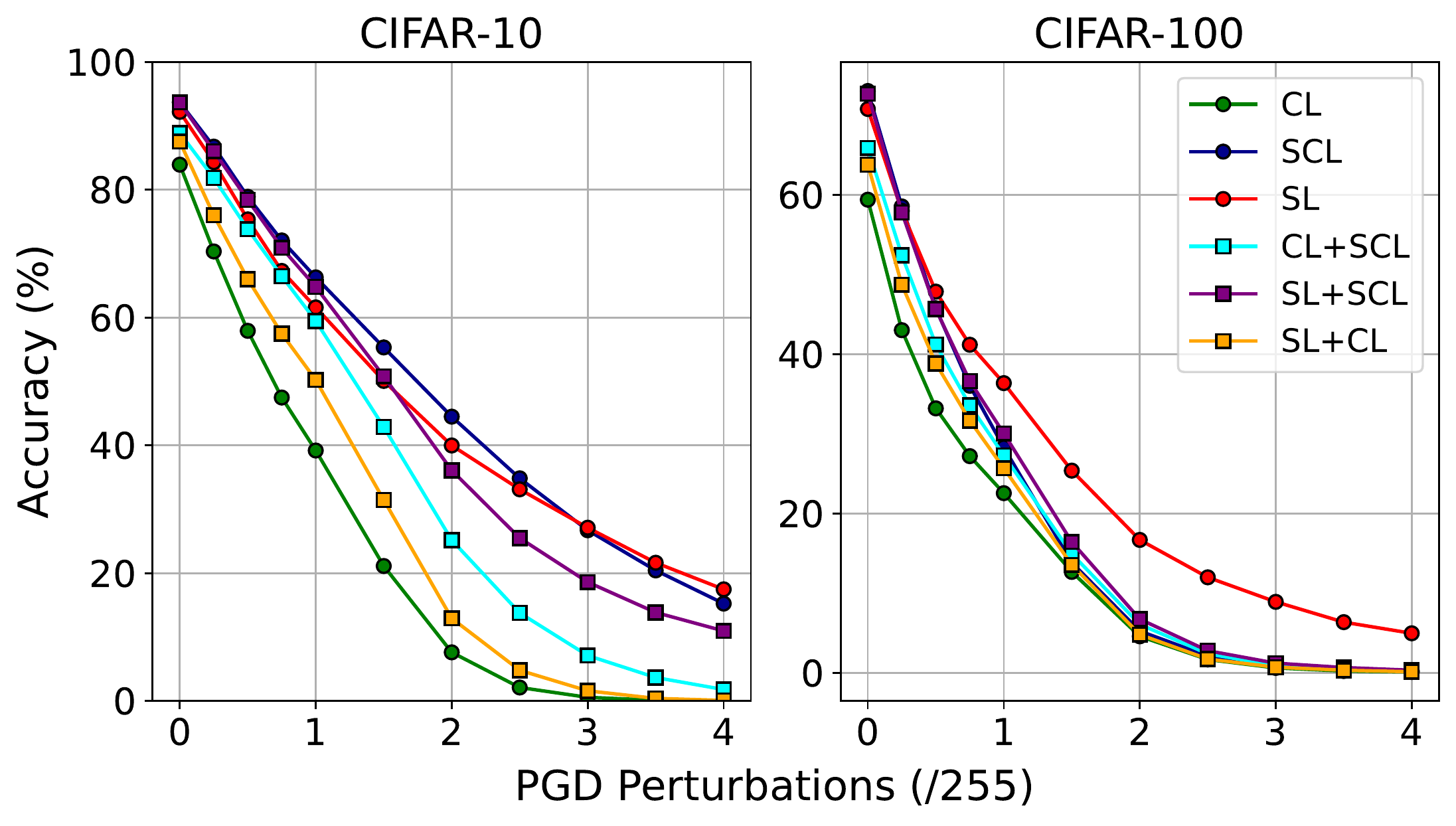}
\caption{\textbf{Incorporating label information into contrastive learning enhances the robustness of the resulting representations.} We compare the test accuracy of different learning schemes on CIFAR10 and CIFAR100 datasets against adversarial examples through standard training settings. Contrastive learning without labels shows lower robustness compared to other learning schemes. We also observed that semi-supervised learning
schemes SL+CL or SCL+CL achieve better robust performance than the CL scheme.}
\label{ST-Semi}
\end{figure}

\section{Experiments}
Our goal is to understand whether there are differences in how contrastive learning learns the representation from data compared to supervised learning from the adversarial perspective. To this end, we conduct extensive experiments to evaluate the robustness of Contrastive Learning (CL), Supervised Contrastive Learning (SCL), and Supervised Learning (SL) under different training scenarios as shown in Table \ref{Ap.pf} on CIFAR-10 and CIFAR-100 image classification benchmarks. In all the subsequent experiments, we train the base encoder and the linear classifier on the same dataset. Our experimental setup is provided in Appendix \ref{ap-Exp-S}. We evaluate the robustness of different scenarios using two threat models: Threat Model-I involves end-to-end attacks generated by cross-entropy loss, where the attacker has complete knowledge of architecture and network parameters in the base encoder and linear classifier. In Threat Model-II, attacks are specifically targeted against the base encoder (see Appendix \ref{ap-threat} for more details).
\subsection{Contrastive Learning: Robustness and Label Impact Analysis through Standard Training} \label{Sec: EX1}
While it is widely known that neural networks trained through standard training are vulnerable to adversarial examples, the degree of vulnerability may differ among models trained using different learning methods. Hence, here we aim to investigate the vulnerability of various learning algorithms through standard training to evaluate their performance. The results as a function of the perturbation size under Threat Model-I are shown in Figure \ref{ST-Semi}. We use 20-step $l_\infty$ Projected Gradient Descent (PGD) \cite{madry2017towards} attacks with different perturbations to generate adversarial attacks during this experiment. In addition to the learning schemes mentioned before, the combination of them including the combination of supervised learning and supervised contrastive learning (denoted SL+SCL), and two other semi-supervised versions, including the combination of contrastive learning with supervised learning (denoted SL+CL) or with supervised contrastive learning (denoted CL+SCL), are investigated. These combinations are designed to answer this question: \textit{Does employing the label information benefit the robustness of contrastive learning representations?} As we can see, contrastive learning without label information is less robust than other learning schemes. We also observed that semi-supervised learning schemes SL+CL or CL+SCL achieve better robust performance than the CL scheme. Appendix \ref{Ap-tsne} visualizes the representations learned by all these learning schemes using t-SNE on the CIFAR-10 dataset.

We have excluded the semi-supervised versions from the subsequent experiments to prevent any potential confusion between the effects of label information and adversarial training.
Appendix \ref{Ap-ST-threat2} provides more
results under Threat Model-II, where the attacks are generated against only the base encoder.

\subsection{Adversarial Training: Comparing Representations}\label{AT-charactristics}
Here, we first compare the performance of different adversarial training scenarios. Subsequently, a set of explanatory tools (e.g., CKA and linear probing) is employed to inspect how adversarial training affects the learned representations in hidden layers.
\subsubsection{Direct Comparison} \label{Sec: EX2}
\begin{figure}[t]
\centering
\includegraphics[scale=0.35]{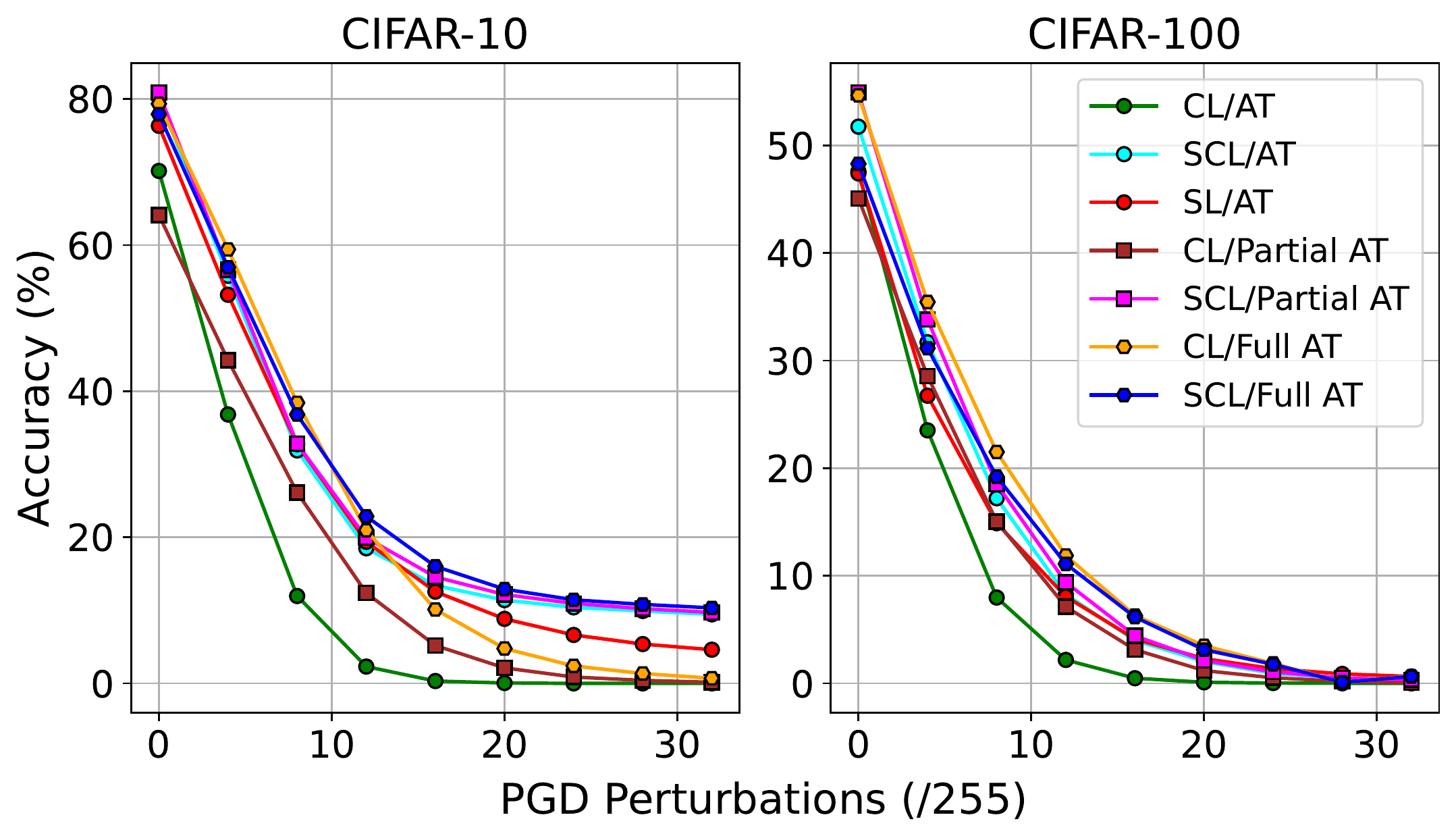}
\caption{\textbf{Full AT successfully integrates label information into the robust network architecture trained via contrastive learning, resulting in a remarkable overall improvement in robustness.}
  We compare the test accuracy of different learning schemes against different PGD attacks on CIFAR10 and CIFAR100 datasets through different adversarial training scenarios. The results demonstrate that Full AT effectively incorporates label information into the adversarially-trained network obtained through the CL scheme. This incorporation leads to improved robustness. Moreover, there is a slight variation in the performance of the SCL under different adversarial training scenarios.}
  \label{AT-PGD}
\end{figure}

This experiment aims to evaluate model robustness under Threat Model-I in the following scenarios: i) training a base encoder using adversarial training, then training the linear classifier separately after freezing the base encoder (AT); ii) training the base encoder by adversarial training, then training the linear classifier separately using adversarial training after freezing the base encoder (Partial AT); iii) training the base encoder by adversarial training, then robustifying the end-to-end model (Full AT). The latter case means that the base encoder parameters are first adversarially trained using $\mathscr{L}_{CL}$ or $\mathscr{L}_{SCL}$, then those parameters are fine-tuned during adversarial training of the linear classifier where adversarial examples are generated using $\mathscr{L}_{CE}$.
The results on the CIFAR-10 and CIFAR100 datasets against 20-step different PGD attacks are shown in
Figure \ref{AT-PGD}. Appendix \ref{Ap-AT-autoattk} also provides the robustness of different scenarios on the CIFAR100 dataset against a state-of-the-art adversarial attack known as Auto-attack \cite{croce2020reliable}. Moreover, Appendix \ref{Ap-non-transferibility} provides more results under Threat Model-II, where the attacks are generated against only the base encoder. The results indicate two main observations: (i) CL under Full AT consistently outperforms other learning schemes in various evaluation scenarios, demonstrating a noticeable improvement in standard accuracy and robustness against adversarial attacks. The results from the previous section shed light on the positive impact of utilizing label information to enhance the robustness of the CL scheme during standard training. It is worth noting that Full AT also effectively incorporates label information into the robust network architecture, leading to a remarkable overall improvement in robustness. This finding is consistent with prior research \cite{zhai2019adversarially,fan2021does} which has investigated the utilization of pseudo labels for unlabeled data, demonstrating their effectiveness in enhancing the adversarial robustness of neural networks. By incorporating label information, the CL scheme benefits from an additional source of valuable supervision, strengthening its defense against adversarial attacks and enhancing overall robustness.
(ii) There is a slight difference in the performance of SCL under AT and Full AT scenarios. This indicates that the representations learned by SCL from the AT scenario are already sufficient to achieve acceptable robustness.


\subsubsection{Comparing characteristics} \label{Sec: EX3}
Previous results raise an important question from a representation learning perspective: what happens in the layers of neural networks when they are adversarially trained? We examine the internal layer representations learned by different robust learning schemes to shed light on this direction.

\begin{figure}[t]
\centering
\includegraphics[scale=0.3]{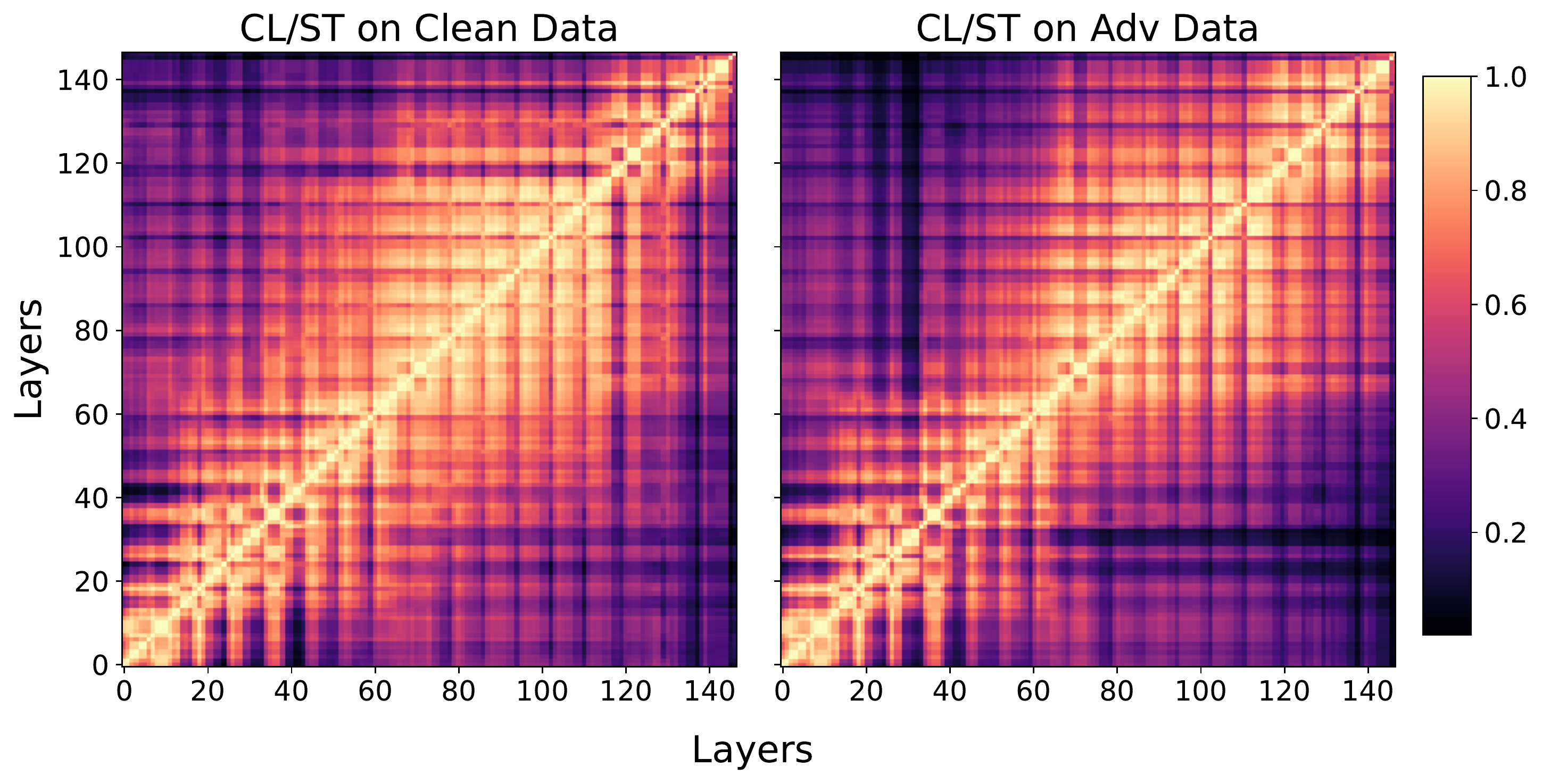}
\caption{\textbf{Standard-trained networks using different learning algorithms yield representations that display notable disparities between adversarial and clean examples.} We compute the similarity of representations across all layer combinations in standard-trained networks that have been trained using the CL scheme, considering both clean and adversarial data.}
  \label{CL-ST-Clean-Adv}
\end{figure}

\vspace{-1em}
\paragraph{Adversarial and clean representations differ significantly in standard-trained networks.} We begin our investigation by using CKA to study the internal representation structure of each model. CKA is a metric that measures the similarity
between two sets of features. To answer how different learning schemes extract representations, we take every pair of layers X and Y within a model learned by different learning schemes and compute their CKA similarity
on clean and adversarial examples. Figures \ref{CL-ST-Clean-Adv} and \ref{All-ST-clean-Adv} show the results as a heatmap for different learning algorithms under standard scenarios on clean and adversarial examples. Notably, we observe distinct differences in the internal representation structure between the three learning schemes: (i) the layers near the end of the network in SCL and SL schemes exhibit lower similarity with other layers compared to CL, and (ii) the dissimilarity between clean and adversarial representations in standard-trained networks highlights their vulnerability to adversarial examples. Previous research \cite{mitrovic2020representation} has assumed that data consists of content and style components, with only the content being relevant for unknown downstream tasks. Additionally, it is assumed that content and style are independent, meaning that style changes do not affect the underlying content.
In contrast, adversarial perturbations are introduced as a specific distribution change in the natural data distribution \cite{zhang2020causal} where they alter the style while preserving the content, making them imperceptible to the human eye. Under these assumptions, the pronounced dissimilarity observed between adversarial and clean examples in standard-trained networks indicates an inability to extract content-related representations consistent across both examples. 

\begin{figure}[t]
\centering
\includegraphics[scale=0.4]{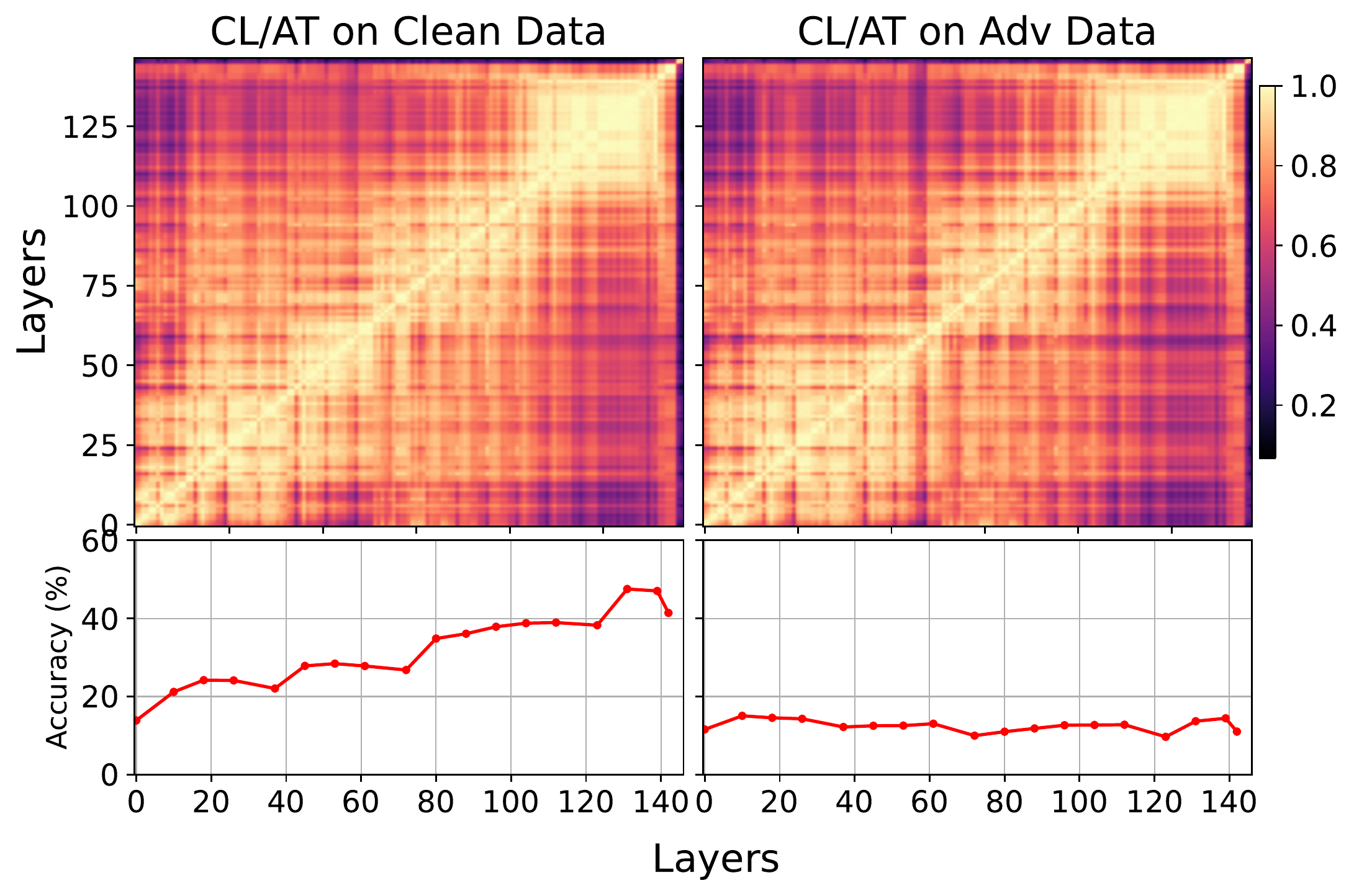}
\caption{\textbf{Regardless of the learning scheme employed, adversarially trained networks exhibit significant similarity between adversarial and clean representations.} We compute the similarity of representations across all layer combinations in adversarially trained networks that have been trained using the CL scheme, considering both clean and adversarial data.}
  \label{CL-AT-Clean-Adv}
\end{figure}
\vspace{-1em}
\paragraph{Adversarial and clean representations exhibit substantial similarity in adversarially trained networks, regardless of the learning schemes used.} We also perform previous comparisons for robust models, taking every pair of layers X and Y within an adversarially trained model robustified by different learning schemes and computing their CKA similarity on clean and adversarial examples. Moreover, we utilize linear probing as a conceptual tool to better understand the dynamics within the neural network and the specific roles played by individual intermediate layers. Figures \ref{CL-AT-Clean-Adv} and \ref{All-Linear_Probing} highlight several observations from the results of the adversarial training experiments: (i) Cross-layer similarities are amplified compared to standard training regardless of learning schemes used. This is evident by the higher degree of brightness in the plots. (ii) In networks trained through adversarial training, the adversarial representations are significantly similar to clean representations. Previous research \cite{jones2022if} has demonstrated that robust training effectively mitigates the impact of adversarial perturbations, resulting in similar representations for clean and adversarial examples in robust networks. However, their studies have only focused on the supervised learning scheme. Our results support and extend these findings by demonstrating that the similarity between clean and adversarial representations holds irrespective of the learning scheme used.
(iii) When comparing the representations obtained from AT and its counterpart, Full AT (see Figure \ref{All-Linear_Probing}), we observe a notable increase in long-range similarities within the CL framework. This enhancement in similarity translates to significant improvements in both standard and adversarial accuracy. Remarkably, Full AT significantly improves overall robustness by incorporating label information into the network. In contrast, the representations learned by SCL and SL under the AT and Full AT scenarios exhibit minor differences. These slight variations in representations result in marginal differences in performance, indicating that the label information utilized in AT already provides sufficient robustness for SCL and SL. Therefore, Full AT does not introduce additional information to enhance SCL and SL's robustness further.

\begin{figure}[t]
\centering
\includegraphics[scale=0.35]{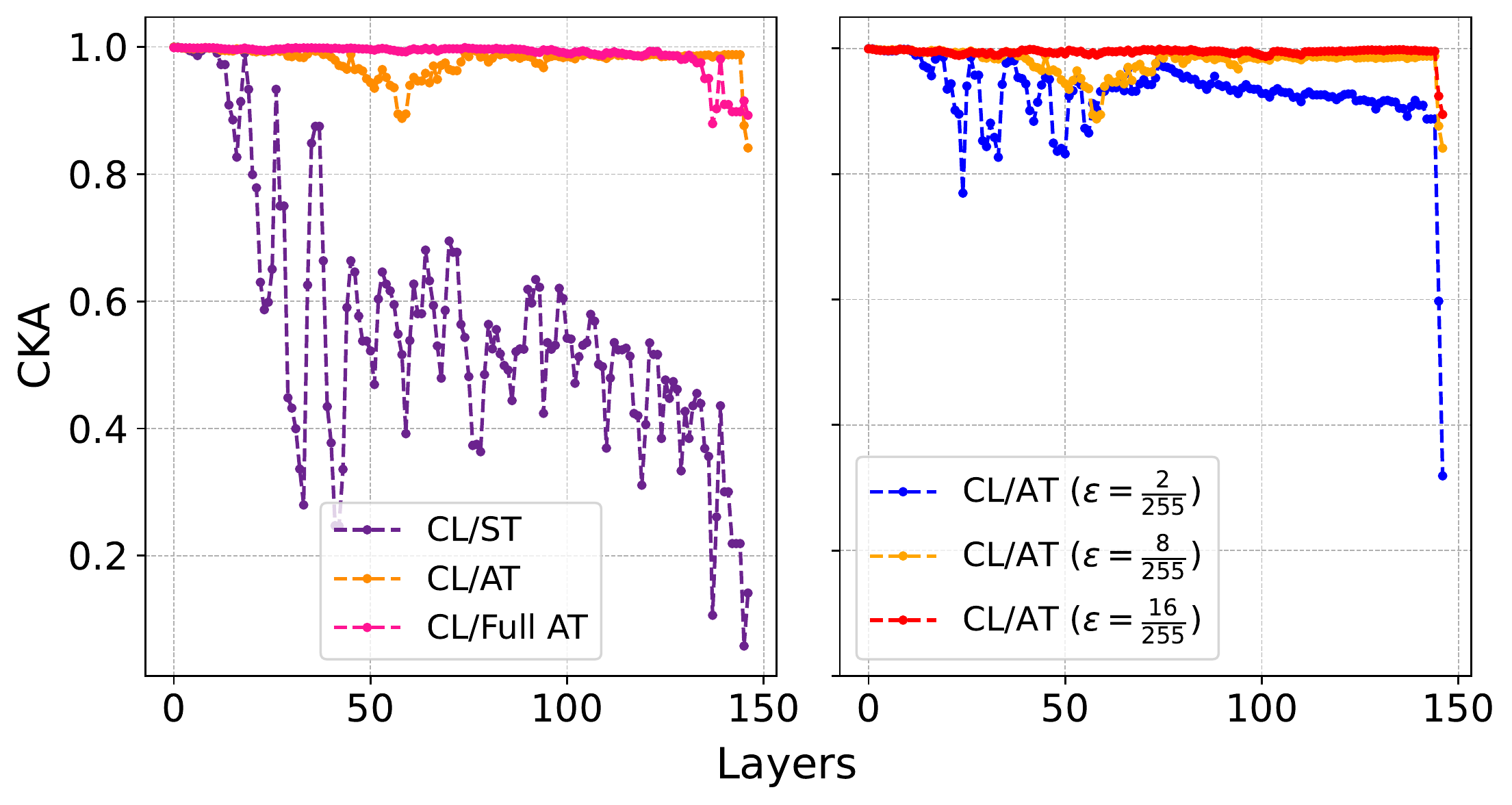}
\caption{\textbf{Increasing the similarity between adversarial and clean representations improves robustness, especially near the end of networks.} Left) Comparing clean and adversarial representations in CL reveals significant dissimilarity in standard-trained networks. Adversarial training reduces this divergence significantly, but a drop in similarity is observed across intermediate layers, affecting its performance (CL/AT). Full AT effectively mitigates the similarity drop, enhancing overall model robustness. Right) During adversarial training, we increased perturbation budgets to vary the strength of adversarial attacks. This led to greater similarity between adversarial and clean representations, especially towards the end of the network.}
  \label{CL-STATFAT-div}
\end{figure}

\vspace{-1em}
\paragraph{Increasing the similarity between adversarial and clean representations, especially near the end of the network, improves robustness.} To gain a deeper understanding of the divergence between adversarial and clean representations, we compare each layer X in a model applied to clean data with its identical counterpart Y in the same model applied to adversarial examples.
The results in the left-hand side of Figure \ref{CL-STATFAT-div} and Figure \ref{All-div-CIFAR100} illustrate that the adversarial representations in the network trained using standard training, exhibit significant dissimilarity from their clean counterparts, particularly towards the end of the network, regardless of the learning scheme used. In contrast, adversarial training significantly reduces the impact of adversarial perturbations, leading to similar representations for both clean and adversarial examples in robust networks. Notably, we observe a similarity drop across intermediate layers of CL/AT, which may explain its lower performance compared to other robust learning schemes. To confirm our explanation, we conducted a similar experiment on CL after Full AT (which significantly improves the robustness) and compared the resulting representations. As shown in the left-hand side of Figure \ref{CL-STATFAT-div}, Full AT reduces the similarity drop, enhancing the model's robustness. To gain further insights into this phenomenon, we conducted an ablation study by varying the strength of adversarial attacks during training through increased perturbation budgets ($\epsilon$). As demonstrated in the right-hand side of Figure \ref{CL-STATFAT-div} and Figure \ref{All-div-diff-epsilon}, we observed that stronger adversarial perturbations led to an enhanced similarity between adversarial representations and their counterpart clean representations, particularly in the later layers of the network. This significant finding confirms the previous results reported in \cite{cianfarani2022understanding} and highlights our novel contribution in extending this observation to contrastive learning schemes. Furthermore, from heatmaps in Figure \ref{AT-diff-epsilon}, we can observe that increasing the strength of adversarial perturbations leads to the long-range similarity between different layers.



\begin{figure}[t]
\centering
\includegraphics[scale=0.3]{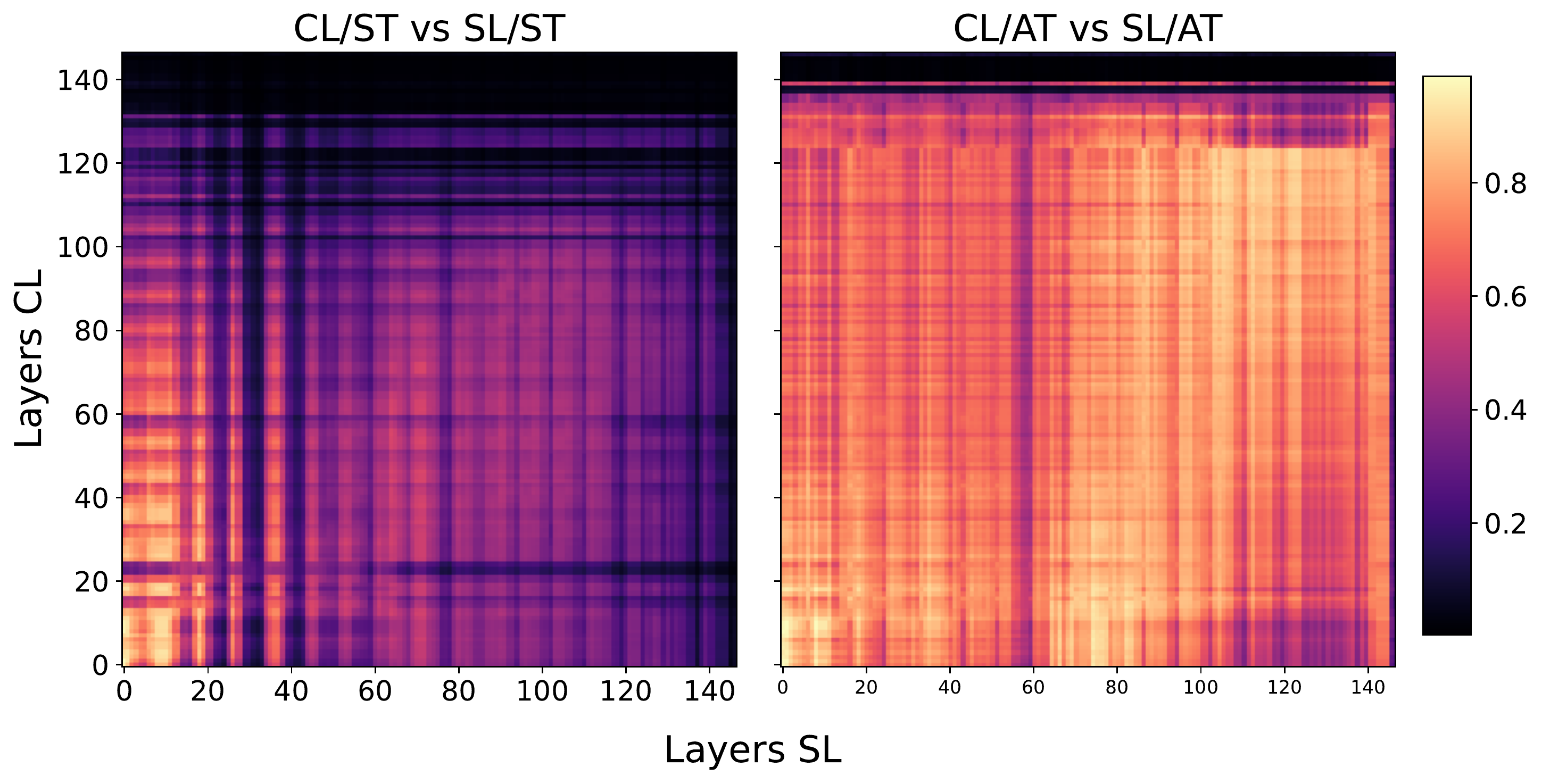}
\caption{\textbf{Adversarial training promotes convergence towards a universal set of representations.} Standard-trained networks exhibit significant dissimilarity in adversarial representations across CL schemes, particularly in higher layers. However, after applying adversarial training, the similarity between layers increases, indicating a shift towards extracting a universal set of representations.}
  \label{CL-SL-ST-AT}
\end{figure}
\vspace{-1em}
\paragraph{Unlike standard training, adversarial training converges toward a universal set of representations, regardless of the learning schemes utilized.}
Here, we perform cross-model comparisons to measure the similarities between all layers X of one model trained using a specific learning scheme and all layers Y of another model trained using a different learning scheme. The left-hand side of
Figures \ref{CL-SL-ST-AT} and \ref{cross-model-all} present the results for models trained using different standard learning schemes. We observed that, except for lower layers, the representations extracted by other layers were highly dissimilar. However, after applying adversarial training (shown on the right-hand side), the similarity between layers from networks trained with different learning schemes notably increased, indicating a tendency towards extracting a universal set of representations.
This finding aligns with a previous study \cite{jones2022if}, which highlighted that robust networks converge towards a universal set of representations regardless of the architecture. Our results extend these observations to contrastive learning schemes, providing empirical evidence for a more general claim. Specifically, our study demonstrates that adversarial training promotes convergence toward a universal set of representations regardless of the learning schemes employed.

\section{Conclusion}
This study compared the robustness of contrastive and supervised contrastive learning with standard supervised learning. Our results demonstrated the benefits of incorporating label information in contrastive learning for enhanced robustness. Adversarial training reduced disparities between adversarial and clean representations, leading to convergence toward a universal set of representations. The increased similarity between adversarial and clean representations improved robustness, especially in deeper layers. These findings offer valuable insights for optimizing robust learning schemes.

 \section*{Acknowledgements}
This work has been supported in part by NSF (Awards 2233873, 2007202, and 2107463), NASA (Award 80NSSC20K1720), and Chameleon Cloud.

\appendix
\clearpage
\numberwithin{figure}{section} 
\numberwithin{table}{section} 
APPENDIX

Below, we provide supplementary information and additional results from the various sections. 

\section{Methodology Details} \label{ap-methodology}
The training process in contrastive and supervised contrastive learning consists of two main phases:

\begin{itemize}
\item\textbf{Pretraining Phase:} In this phase, the primary objective is to train the base encoder parameters $\bm{\theta}{b}$ by minimizing a self-supervised loss function $\mathscr{L}{p}(\bm{\theta}{b},\bm{\theta}{ph})$ using a dataset $\mathscr{D}{p}$. The parameters $\bm{\theta}{ph}$ correspond to the projection head, which maps the output of the base encoder into a lower-dimensional latent space where the self-supervised loss $\mathscr{L}_{p}$ is applied.
\\
\item\textbf{Supervised Fine-tuning Phase:} In this phase, the main objective is to train the linear classifier parameters $\bm{\theta}_{c}$ by minimizing a supervised loss function $\mathscr{L}f(\bm{\theta}{c})$ using a labeled dataset $\mathscr{D}_f$. The purpose of the linear classifier is to map the representations obtained during the pretraining phase to the labeled space, utilizing the cross-entropy loss $\mathscr{L}_f$. 
\end{itemize}
In the \textit{standard training of pretraining phase}, $\mathscr{L}_{p}$ is given by $\mathscr{L}_{CL}(\bm{\theta}_{b},\bm{\theta}_{ph};\bm{x}^\prime,\bm{x}^{\prime\prime})$ and $\mathscr{L}_{SCL}(\bm{\theta}_{b},\bm{\theta}_{ph};\bm{x}^\prime,\bm{x}^{\prime\prime},\bm{y})$ introduced in \cite{chen2020simple} and \cite{khosla2020supervised} for the contrastive and supervised contrastive schemes, respectively. Here the $\bm{x}^\prime$ and $\bm{x}^{\prime\prime}$ are two transformed views of the same minibatch and label vector $\bm{y}$ is leveraged in  $\mathscr{L}_{SCL}$ to avoid false positive pair selection, as explained in \cite{khosla2020supervised}. 

In the scenario of \textit{robust pretraining or adversarial training of the representations}, we adopt a loss function $\mathscr{L}_{p}$ inspired by Adversarial Contrastive Learning \cite{jiang2020robust}. This loss function is formulated as a linear combination of two terms, as defined below:
\\
\begin{equation}
\begin{array}[b]{rclcl}
\mathscr{L}_{p}(\bm{\theta}_{b},\bm{\theta}_{ph}) &=& \alpha\, \mathscr{L}_{CL}(\bm{\theta}_{b},\bm{\theta}_{ph};\bm{x}^\prime,\bm{x}^{\prime\prime})\\
&+& \beta\, \mathscr{L}_{CL}(\bm{\theta}_{b},\bm{\theta}_{ph};\bm{x},\bm{x}_{adv}) 
\end{array}
\end{equation}
and
\\
\begin{equation}
\begin{array}[b]{rclcl}
\mathscr{L}_{p}(\bm{\theta}_{b},\bm{\theta}_{ph}) &=& \alpha\, \mathscr{L}_{SCL}(\bm{\theta}_{b},\bm{\theta}_{ph};\bm{x}^\prime,\bm{x}^{\prime\prime},\bm{y}) \\&+&
\beta\, \mathscr{L}_{SCL}(\bm{\theta}_{b},\bm{\theta}_{ph};\bm{x},\bm{x}_{adv},\bm{y}) 
\end{array}
\end{equation}
for the contrastive and supervised contrastive schemes, respectively. In these equations, $\bm{x}^\prime$ and $\bm{x}^{\prime\prime}$ are the transformed views of $\bm{x}$, while $\bm{x}_{adv}$ is the PGD attack generated by maximizing the associated loss function iteratively over each given minibatch $\bm{x}$ as follows:
\\
\begin{equation}
\bm{x}^{t+1} = \Pi_{\bm{x}+\bm{S}}{(\bm{x}^t+\alpha} sgn(\nabla_{\bm{x}} \mathscr{L}_{CL}(\bm{\theta}_{b},\bm{\theta}_{ph};\bm{x},\bm{x}_{adv}))
\end{equation}
\\
and
\\
\begin{equation}
\small
\bm{x}^{t+1} = \Pi_{\bm{x}+\bm{S}}{(\bm{x}^t+\alpha} sgn(\nabla_{\bm{x}} \mathscr{L}_{SCL}(\bm{\theta}_{b},\bm{\theta}_{ph};\bm{x},\bm{x}_{adv},\bm{y}))
\end{equation}
\\
for the contrastive and supervised contrastive schemes, respectively. Here, $\Pi_{\bm{x}+\bm{S}}$ denotes projecting perturbations into the set of allowed perturbations $\bm{S}$
and $\alpha$ is the step size.
In this setup, the minimization of $\mathscr{L}_{p}$ corresponds to the simultaneous minimization of each term weighted by coefficients of $\alpha$ and $\beta$. In this study, we take $\alpha=\beta$ to avoid unjustified prioritization between the transformed and adversarial terms. 

We consider two alternatives for the \textit{robust training of the fine-tuning phase}. In the \textit{partial adversarial training}, we \underline{only} update the linear classifier parameters $\theta_c$ by minimizing the loss function,
\\
\begin{equation}
\mathscr{L}_f(\bm{\theta}_{c}) = \alpha\, \mathscr{L}_{CE}(\bm{\theta}_{c};\bm{x},\bm{y}) + \beta\, \mathscr{L}_{CE}(\bm{\theta}_{c};\bm{x}_{adv},\bm{y}) 
\end{equation}
\\
As the second alternative, \textit{full adversarial fine-tuning} utilizes the following loss function,
\\
\begin{equation}
\begin{array}[b]{rclcl}
\mathscr{L}_f(\bm{\theta}_{b},\bm{\theta}_{c}) &=& \alpha\, \mathscr{L}_{CE}(\bm{\theta}_{b},\bm{\theta}_{c};\bm{x},\bm{y})\\
&+& \beta\, \mathscr{L}_{CE}(\bm{\theta}_{b},\bm{\theta}_{c};\bm{x}_{adv},\bm{y}) 
\end{array}
\end{equation}
\\
to \underline{readjust} the base encoder parameters $\bm{\theta}_b$ and train the linear classifier. In these equations, $\bm{x}_{adv}$ is the PGD attack generated by maximizing the cross-entropy loss iteratively over each given minibatch $\bm{x}$. Here we take $\alpha=\beta$ in parallel with the pretraining phase. 

We investigate both the standard and robust training approaches of the aforementioned training phases to compare the adversarial robustness among various learning schemes. 
A summary of all the training combinations studied, including various scenarios of training phases in contrastive and supervised contrastive learning schemes, can be found in Table \ref{Ap.pf}.

\begin{table}[h]
\centering
\small
\caption{Summary of the training scenarios.}
\resizebox{0.49\textwidth}{!}{%
\begin{tabular}{ccc}
\toprule
\textbf{Scenarios} & \textbf{Pretraining Phase} & \textbf{Finetuning Phase} \\
\midrule
ST & Standard Training & Standard Training \\
 & & (with fixed $\bm{\theta}_b$) \\
\addlinespace
AT & Adversarial Training & Standard Training \\
 & & (with fixed $\bm{\theta}_b$) \\
\addlinespace
Partial-AT & Adversarial Training & Partial Adversarial Training \\
 & & (with fixed $\bm{\theta}_b$) \\
\addlinespace
Full-AT & Adversarial Training & Full Adversarial Training \\
\bottomrule
\end{tabular}
}
\label{Ap.pf}
\end{table}

\section{Experiment Setup}\label{ap-Exp-S}
Our experiment setup is similar to that used in \cite{chen2020simple} for SimCLR and \cite{khosla2020supervised} for SupCon, which are prominent works in contrastive learning. We use ResNet-50 as the base encoder for all scenarios and a two-layers MLP network as the projection head. The loss is optimized using the Adam optimizer with a learning rate of 0.0003. We train each model for 200 epochs using a mini-batch size of 128 for standard and 256 for adversarial scenarios. In all adversarial training scenarios, the adversarial perturbations are generated using a 5-step Projected Gradient Descent (PGD) attack under the $l_\infty$ norm with a maximum perturbation limit of $\epsilon = 8/255$, unless a specific value of $\epsilon$ is specified. The models are evaluated against PGD attacks and state-of-the-art Auto-attacks \cite{croce2020reliable} at the test time. We report the top-1 test accuracy for all scenarios
to evaluate the mentioned scenarios.

\subsection{Threat Models} \label{ap-threat}
To evaluate the robustness of each scenario, we consider two different threat models:
\begin{itemize}[noitemsep,topsep=0pt,leftmargin=*]
\item Threat Model-I (end-to-end attack generated by $\mathscr{L}_{CE}$): In this threat model, the attacker has complete knowledge of architecture and network parameters in the base encoder and linear classifier. This excludes any knowledge about the projection head utilized in the pretraining phase. The attacks are generated end-to-end through the utilization of the cross-entropy loss. 
\item Threat Model-II (attack generated against base encoder by $\mathscr{L}_{p}$): In Threat Model-II, the attacker possesses complete knowledge of all components in the pretraining phase, including the architecture, network parameters, loss function, and training dataset. 
\end{itemize}

\section{t-SNE Visualization of Learning Schemes under Standard Training Scenario}\label{Ap-tsne}

\begin{figure}[h]
\centering
\includegraphics[scale=0.2]{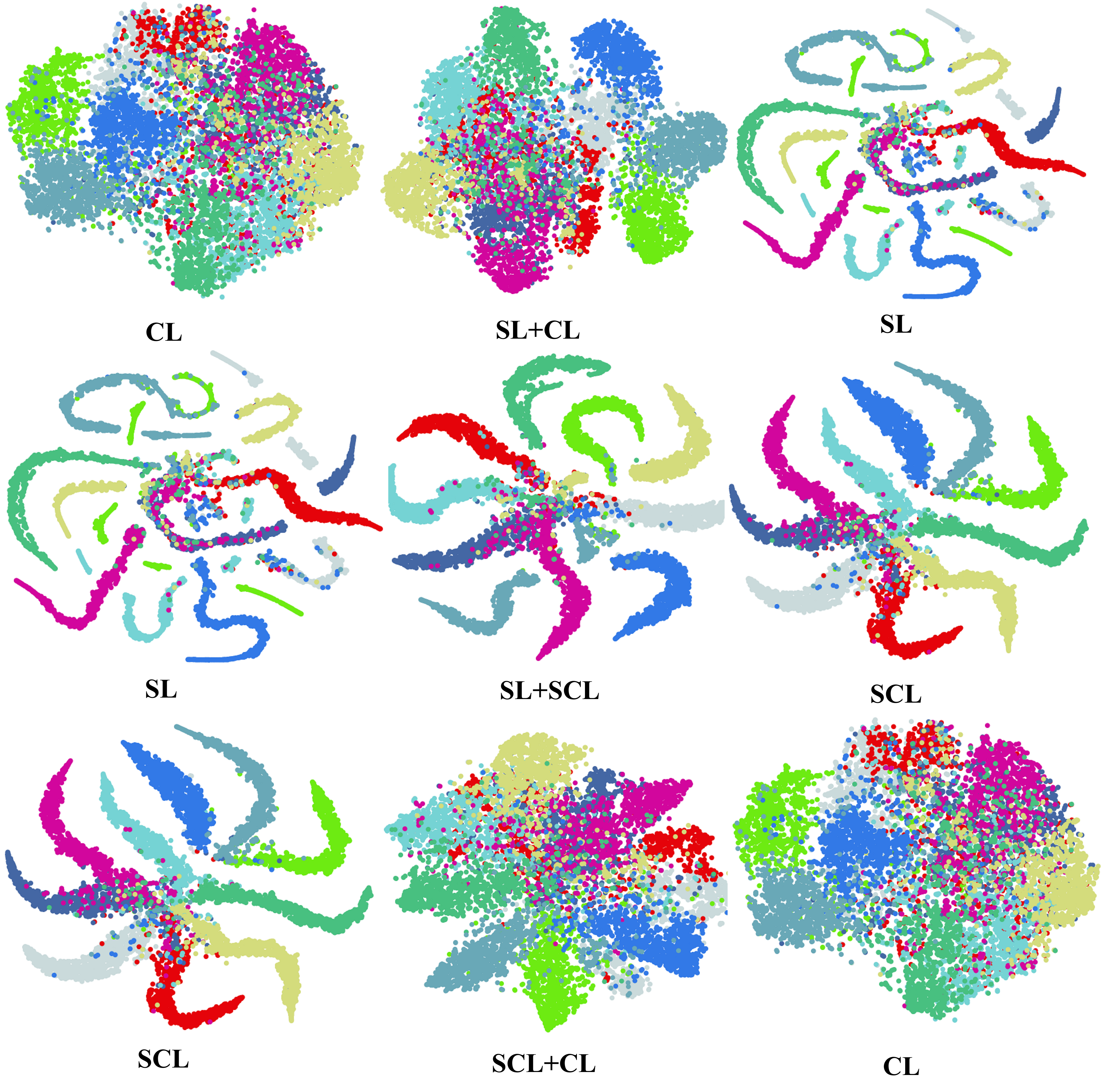}
\caption{\textbf{Semi-supervised learning schemes (SL+CL and SCL+CL) separate classes more clearly than contrastive learning (CL) schemes.} We visualize the representations learned by different learning schemes (CL, SCL, SL, SL+SCL, SL+CL, and SCL+CL) using t-SNE on the CIFAR-10 dataset. The results show that in the ST scenario, both SCL and SL exhibit clearer class boundaries compared to CL. Furthermore, incorporating label information in the semi-supervised learning schemes (SL+CL and SCL+CL) enhances the separation of classes, indicating increased robustness against adversarial perturbations. }
\label{tsne}
\end{figure}

Figure \ref{tsne} visualizes the representations learned by CL, SCL, SL, SL+SCL, SL+CL, and CL+SCL learning schemes using t-SNE on the CIFAR-10 dataset. Here, we used labels to color the markers corresponding to each data point. The results depicted in the ST scenario clearly show much clearer class boundaries in the SCL and SL compared to the CL scheme. Furthermore, we can observe that employing some label information in both semi-supervised learning schemes, SL+CL and CL+SCL, can lead to separate classes more clearly. This suggests that semi-supervised learning versions make it difficult for an adversary to successfully perturb an image, resulting in a more robust prediction.

\section{Robustness through Standard Training under Threat Model-II}\label{Ap-ST-threat2}
Threat Model-II is not applicable for the supervised learning scheme, as the base encoder and linear classifier are trained together end-to-end. Table \ref{ST-ST} reports the results on CIFAR-10 and CIFAR-100 datasets against different 40-step PGD attacks. From the results, we can observe that the models trained using $\mathscr{L}_{SCL}$ are more robust compared to $\mathscr{L}_{CL}$ where the attacks are generated against the base encoder. This suggests that false negative pair selection in self-supervised contrastive learning leads to making the model less robust which is aligned with the results reported in \cite{gupta2023contrastive}.

\begin{table}[h]
\centering
\small
\caption{\textbf{SCL is more robust than CL scheme against adversarial attacks generated against base encoder.} The performance of contrastive learning schemes, including CL and SCL, in the ST scenario evaluated under Threat Model-II (attack generated against base encoder). The best performance is highlighted in bold. Threat Model-II is not applicable to the SL scheme, as the base encoder and linear classifier are trained together end-to-end.}
\label{ST-ST}
\resizebox{0.49\textwidth}{!}{%
\begin{tabular}{cccccc}
\toprule
\textbf{Models} & \textbf{Dataset} & \textbf{Standard Training} & \textbf{PGD (4/255)} & \textbf{PGD (4/255)} & \textbf{PGD (16/255)} \\
\midrule
\textbf{SCL} & \multirow{2}{*}{\textbf{CIFAR10}} & 93.56 & \textbf{30.3} & \textbf{12.9} & \textbf{10.06} \\
\addlinespace
\textbf{CL} & & 84.27 & 19.58 & 9.61 & 7.19 \\
\addlinespace
\textbf{SCL} & \multirow{2}{*}{\textbf{CIFAR-100}} & 73.38 & \textbf{7.87} & \textbf{3.28} & \textbf{2.16} \\
\addlinespace
\textbf{CL} & & 60.28 & 5.52 & 0.94 & 0.23 \\
\bottomrule
\end{tabular}%
}
\end{table}

\section{Performance of Different Adversarial Training Scenarios against a Range of Auto-Attacks}\label{Ap-AT-autoattk}
\begin{figure}[H]
\centering
\includegraphics[scale=0.35]{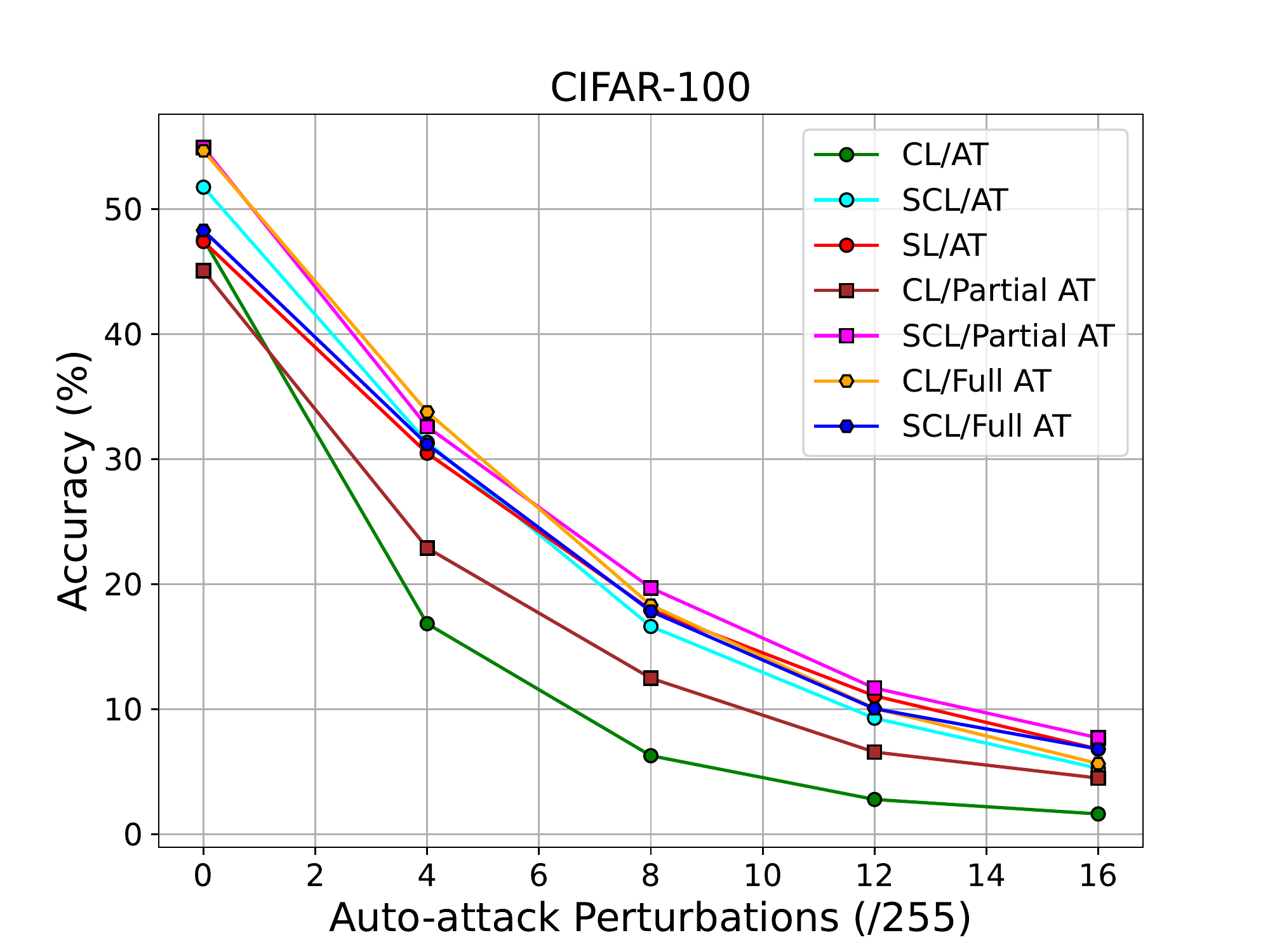}
\caption{\textbf{Full AT effectively incorporates label information into the robust network trained via contrastive learning, significantly improving overall robustness.} The comparative analysis of different learning schemes' test accuracy against various Auto-attacks on the CIFAR-100 dataset demonstrates the efficacy of Full AT in enhancing the robustness of the CL-based network by integrating label information. This integration leads to improved robustness against adversarial attacks. Additionally, slight variations in the performance of SCL are observed under different adversarial training scenarios.}
\label{AT-autoattack}
\end{figure}

Here, we evaluate the robustness of different scenarios on the CIFAR100 dataset against different Auto-attacks. The results are shown in Figure \ref{AT-autoattack}. The comparative analysis provides evidence of Full AT being effective in improving the robustness of the CL-based network. This improvement is achieved through the integration of label information, resulting in enhanced robustness against adversarial attacks. Furthermore, slight variations in the performance of SCL are observed across different scenarios of adversarial training.

\section{Robustness through Adversarial Training under Threat Model-II (Non-Transferability of Cross-Task Robustness)}\label{Ap-non-transferibility}

In contrastive learning, any objective that utilizes the learned representations of the base encoder is referred to as a downstream task.
In this experiment, we compare model robustness where adversarial training is applied only to the base encoder in contrastive and supervised contrastive learning schemes. Table \ref{tb2} shows the performance of the models on CIFAR-10 and CIFAR-100 datasets against different 40-step PGD attacks. Compared to Threat Model-I, our findings in Threat Model-II indicate that the robust model achieved by applying adversarial training solely to the base encoder is not transferable to the downstream task. This failure in transferring robustness across the tasks, known as the cross-task robustness transferability challenge, has also been reported in \cite{fan2021does}.

\begin{table*}[h]
\centering
\caption{\textbf{The robustness achieved through adversarial training solely applied to the base encoder does not transfer effectively to the downstream task.} The performance of contrastive learning schemes, including CL and SCL in AT scenario, is compared to the baseline adversarially trained SL scheme in terms of top-1 accuracy on CIFAR-10 and CIFAR-100 datasets. The models are evaluated under Threat Model-I (end-to-end attack generated by cross-entropy loss) and threat model-II (attack generated against base encoder). The effectiveness of AT in enhancing the robustness of CL and SCL under Threat Model-II is evident. However, the results reveal that these robust models lose their robustness when subjected to end-to-end attacks or under Threat Model-I.}
\label{tb2}
\renewcommand{\arraystretch}{1.3}
\resizebox{0.95\textwidth}{!}{%
\begin{tabular}{cccccccccc}
\toprule
\textbf{Models} & \textbf{Datasets} & \textbf{Standard Training} & \textbf{Adversarial Training} & \multicolumn{3}{c}{\textbf{End-to-End Attack Generated by Cross-Entropy Loss}} & \multicolumn{3}{c}{\textbf{Attack Generated Against Base Encoder}} \\
\addlinespace
\cline{5-7} \cline{8-10}
& & \textbf{Clean} & \textbf{Clean} & \textbf{PGD (4/255)} & \textbf{PGD (8/255)} & \textbf{PGD (16/255)} & \textbf{PGD (4/255)} & \textbf{PGD (8/255)} & \textbf{PGD (16/255)} \\
\addlinespace
\midrule
\textbf{SL} & & 91.73 & 76.33 & 53.2 & 32.52 & 10.5 & NA & NA & NA \\
\addlinespace
\textbf{SCL} & \textbf{CIFAR-10} & 93.56 & 80.22 & 11.99 & 55.83 & 31.45 & \textbf{59.36} & \textbf{62.25} & \textbf{59.38} \\
\addlinespace
\textbf{CL} & & 84.27 & 70.13 & 36.86 & 11.91 & 0.26 & \textbf{69.1} & \textbf{65.88} & \textbf{49.8} \\
\addlinespace
\textbf{SL} & & 69.56 & 47.4 & 26.68 & 14.8 & 3.8 & NA & NA & NA \\
\addlinespace
\textbf{SCL} & \textbf{CIFAR-100} & 73.38 & 51.8 & 31.7 & 17.11 & 3.65 & \textbf{37.09} & \textbf{30.02} & \textbf{24.5} \\
\addlinespace
\textbf{CL} & & 60.28 & 47.58 & 23.58 & 7.83 & 0.43 & \textbf{46.18} & \textbf{42.44} & \textbf{26.9} \\
\addlinespace
\bottomrule
\end{tabular}%
}
\end{table*}

\section{Representation Structure of Different Learning Schemes}\label{Ap-ST-Clean-Adv}
In this section, we provide all the figures related to the section \ref{Sec: EX3} in the main body.

\begin{figure}[ht]
  \vskip 0.2in
  \centering
 \includegraphics[width=0.49\textwidth]{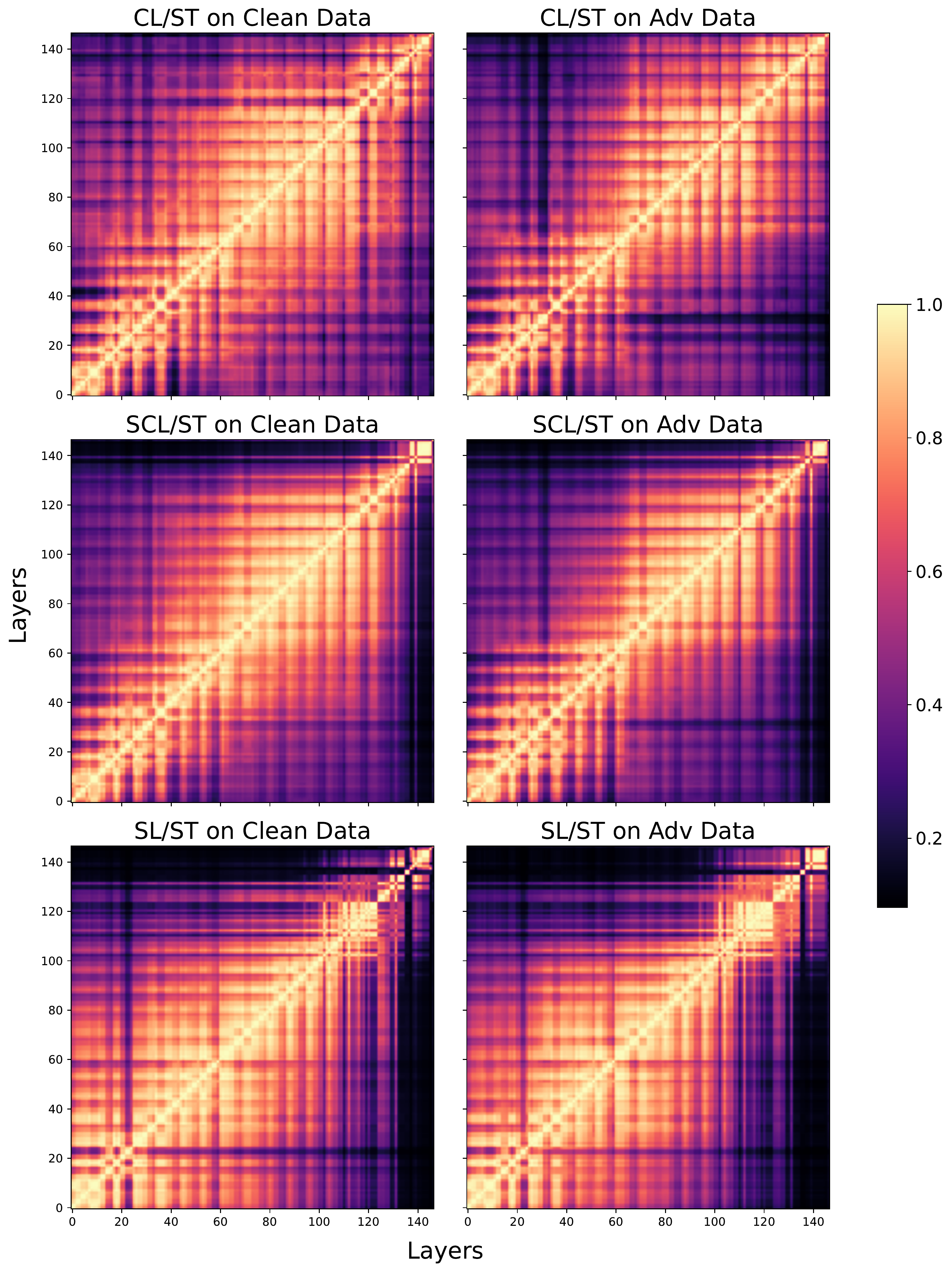}
\caption{\textbf{The representations obtained from standard-trained networks exhibit significant differences between adversarial and clean examples, regardless of the learning algorithm utilized.} We compute the similarity of representations across all layer combinations in standard-trained networks that have been trained using different learning schemes, considering both clean and adversarial data. The three learning schemes (SCL, SL, and CL) have noticeable differences in their internal representation structures (the first column). CL demonstrates more consistent representations throughout the network when compared to SCL and SL. Moreover, standard-trained networks exhibit substantial dissimilarity between clean and adversarial representations (the first column vs. the second one).}
\label{All-ST-clean-Adv}
  \vskip -0.2in
\end{figure}

\begin{figure*}[ht]
  \vskip 0.2in
  \centering
 \includegraphics[width=0.7\textwidth]{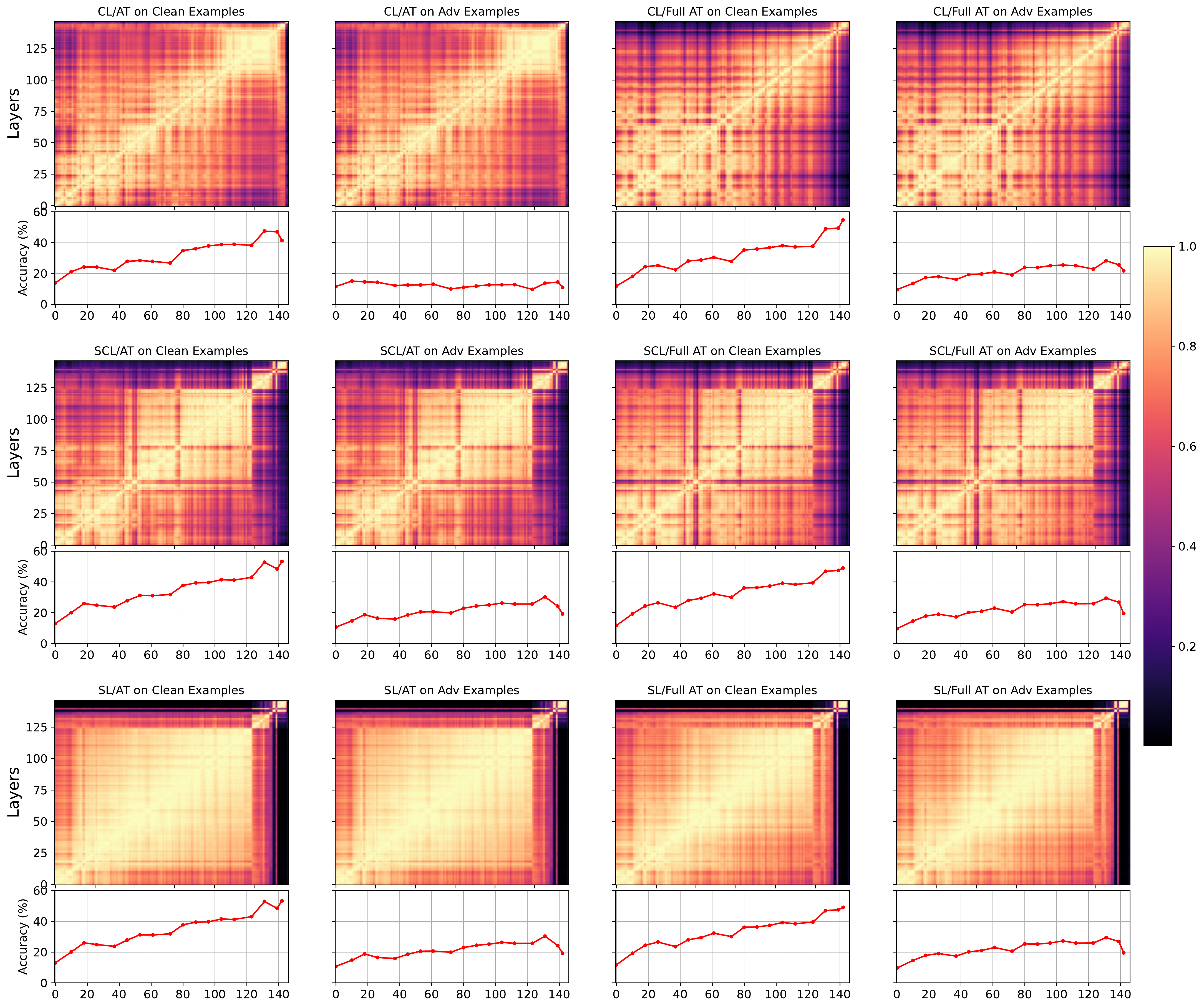}
 \caption{\textbf{The similarity between adversarial and clean representations is substantial in adversarially trained networks, regardless of the learning scheme used.} We analyze robust models by comparing layer pairs within different learning schemes and calculating their CKA similarity on clean and adversarial examples. Linear probing is employed to gain insights into the network dynamics and the roles of intermediate layers. The results demonstrate amplified cross-layer similarities compared to standard training, indicated by higher brightness levels in the plots. Additionally, networks trained through adversarial training exhibit significant similarities between adversarial and clean representations. Moreover, upon comparing the representations obtained from AT and its counterpart Full AT, we observe a significant enhancement in long-range similarities within CL. This improvement in similarity leads to substantial improvements in both standard and adversarial accuracy. In contrast, the representations learned by SCL and SL under AT and Full AT scenarios exhibit slight differences, resulting in minor variations in their performance.}
\label{All-Linear_Probing}
  \vskip -0.2in
\end{figure*}

\begin{figure*}[ht]
  \vskip 0.2in
  \centering
 \includegraphics[width=0.73\textwidth]{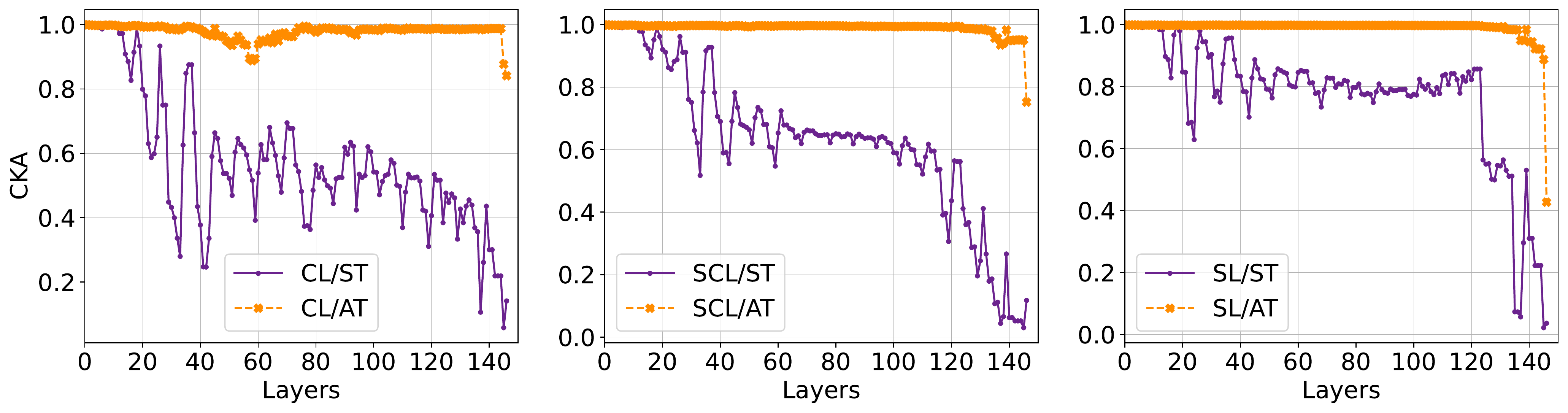}
\caption{\textbf{Contrasting adversarial representations with their clean counterparts.} Comparing clean and adversarial representations in different layers of the model reveals significant dissimilarity in standard-trained networks. Adversarial training reduces this divergence, leading to similar representations for clean and adversarial examples in robust networks. However, there is a drop in similarity across intermediate layers for CL/AT, which may explain its lower performance compared to other robust learning schemes.}
  \label{All-div-CIFAR100}
  \vskip -0.2in
\end{figure*}


\begin{figure*}[ht]
  \vskip 0.2in
  \centering
 \includegraphics[width=0.73\textwidth]{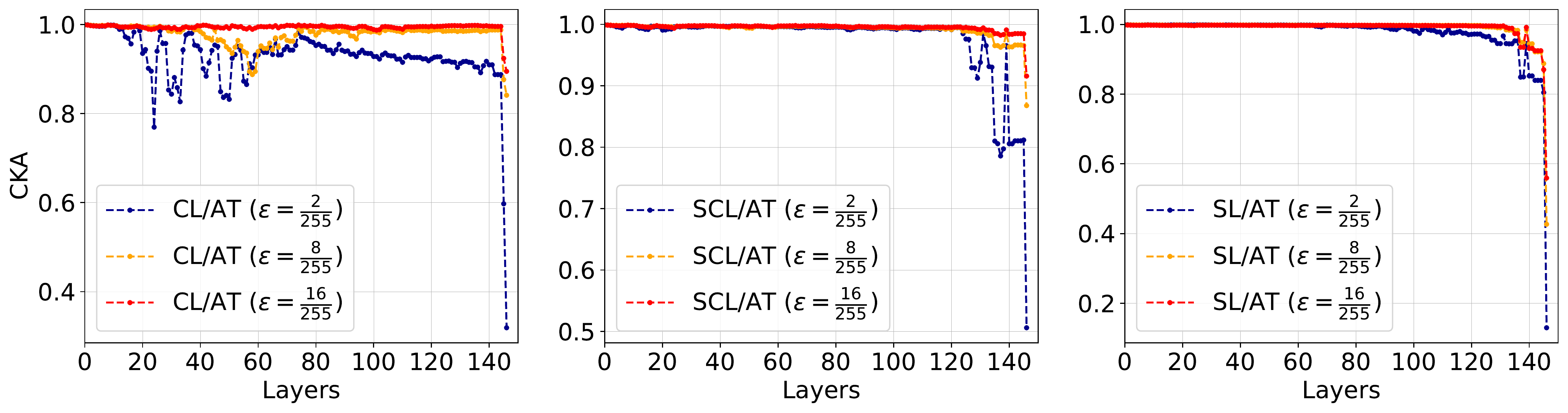}
\caption{\textbf{Increasing the similarity between adversarial and clean representations improves robustness, especially near the end of networks.} Improving the similarity between adversarial and clean representations enhances robustness. During adversarial training, we increased perturbation budgets to vary the strength of adversarial attacks. This led to greater similarity between adversarial and clean representations, especially towards the end of the network.}
  \label{All-div-diff-epsilon}
  \vskip -0.2in
\end{figure*}

\begin{figure*}[ht]
  \vskip 0.2in
  \centering
 \includegraphics[width=0.73\textwidth]{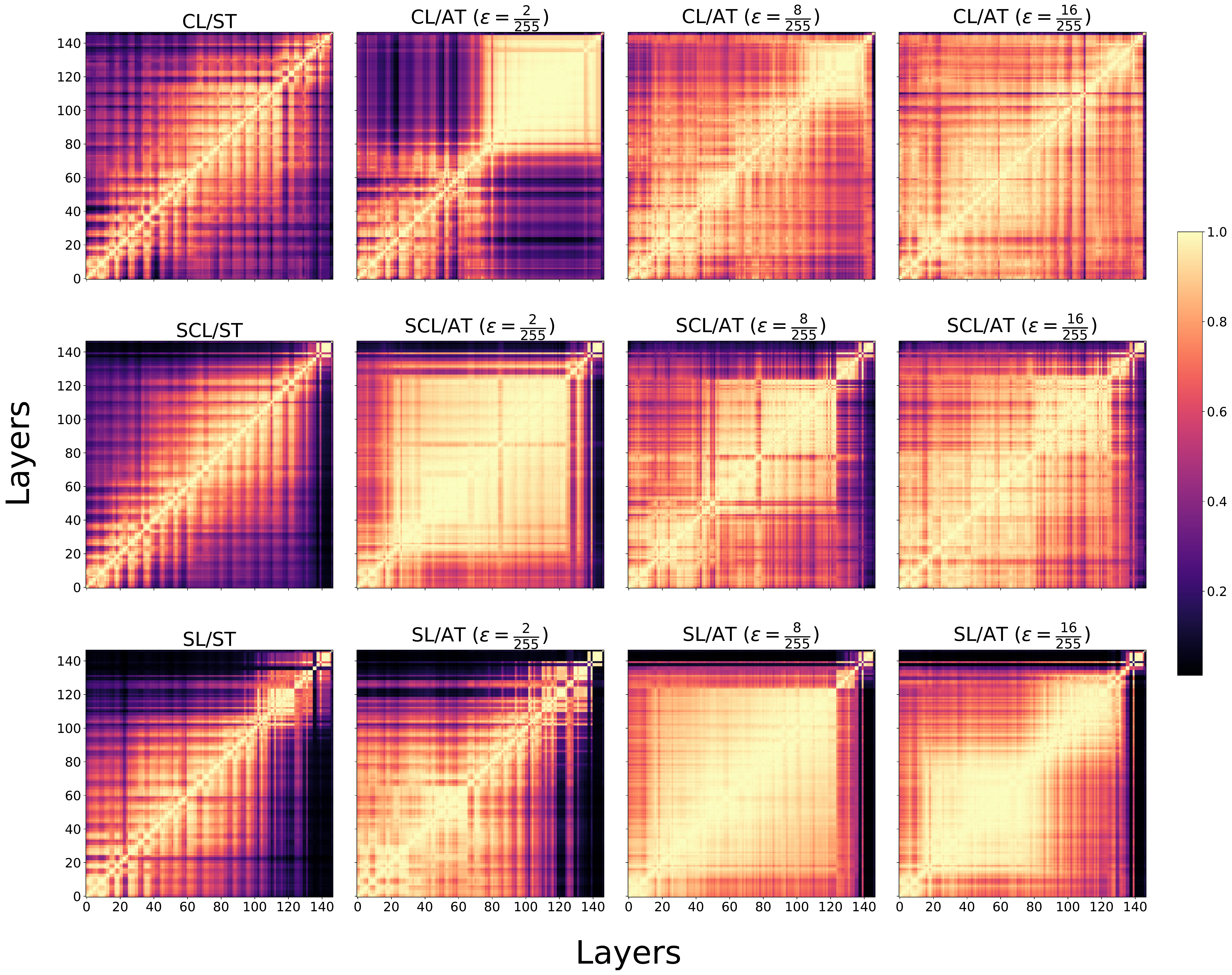}
 \caption{\textbf{Adversarial training promotes the emergence of long-range similarities between layers, regardless of the specific learning scheme employed.} We vary the strength of adversarial attacks during training. We observe that increasing the strength of adversarial perturbations leads to a consistent presence of long-range similarity between layers, independent of the learning scheme used. }
\label{AT-diff-epsilon}
  \vskip -0.2in
\end{figure*}

\begin{figure*}[ht]
  \vskip 0.2in
  \centering
 \includegraphics[width=0.49\textwidth]{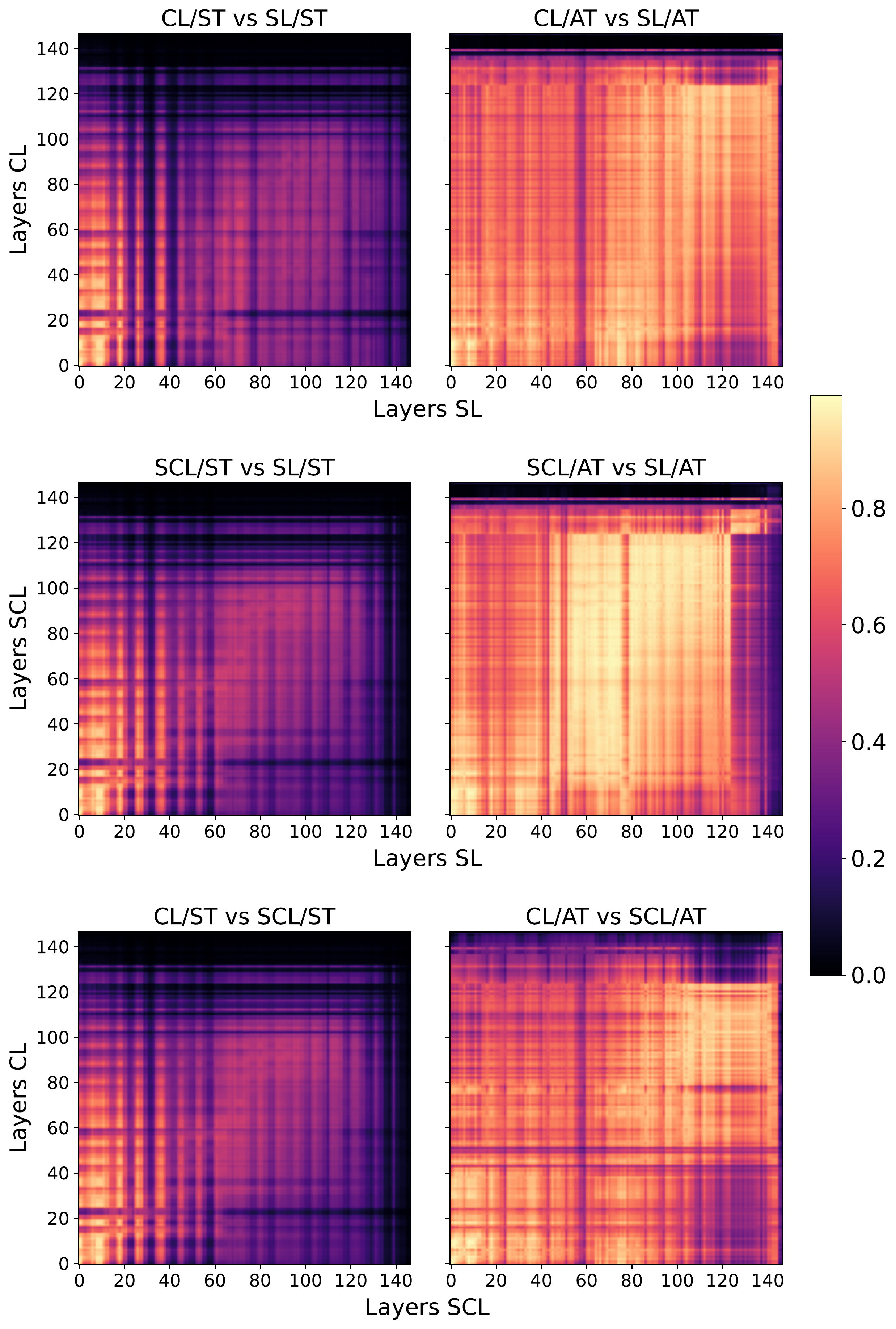}
  \caption{\textbf{Unlike standard-trained networks, the ones trained through adversarial training show significant similarity in adversarial representations across different learning schemes. } The cross-model CKA heatmap between standard-trained networks trained using different learning schemes highlights that these schemes extract distinct adversarial representations, particularly in a large number of higher layers within the network.
  Cross-model comparisons demonstrate that, after applying adversarial training, the similarity between layers from different learning schemes increases, suggesting a shift towards extracting a universal set of representations.}
  \label{cross-model-all}
  \vskip -0.2in
\end{figure*}

\end{document}